\documentclass[letterpaper, 10 pt, conference]{ieeeconf}  

\IEEEoverridecommandlockouts                              

\usepackage{amsmath} 
\usepackage{amssymb}  
\usepackage{graphicx}
\usepackage{capt-of}
\usepackage{bm}
\usepackage{caption, subcaption}
\usepackage{subcaption}

\makeatletter
\let\NAT@parse\undefined
\makeatother
\usepackage[hidelinks]{hyperref}
\usepackage[dvipsnames]{xcolor}
\usepackage{dsfont}

\usepackage{outlines}
\usepackage{amsfonts}
\usepackage{booktabs}
\usepackage{multicol}
\usepackage{siunitx}
\usepackage{tikz,graphics,float,epsf}
\usepackage{tabularx}
\usepackage{times}
\usepackage{bbm}
\usetikzlibrary{calc}
\usepackage{xspace}
\usepackage{layouts}

\usepackage[normalem]{ulem}

\definecolor{dark_green}{rgb}{0.18, 0.55, 0.35} 
\definecolor{dark_orange}{rgb}{0.8, 0.45, 0.0}

\newcounter{nodecount}
\newcommand\tabnode[1]{\addtocounter{nodecount}{1} \tikz \node (\arabic{nodecount}) {#1};}

\tikzstyle{every picture}+=[remember picture,baseline]
\tikzstyle{every node}+=[inner sep=0pt,anchor=base,
text depth=.25ex,outer sep=1.5pt]
\tikzstyle{every path}+=[thick, dashed, rounded corners]

\title{\LARGE \bf
Learning Generalizable Tool Use\\ with Non-rigid Grasp-pose Registration
}

\author{Malte Mosbach and Sven Behnke
\thanks{Both authors are with the Autonomous Intelligent Systems group, University of Bonn, Germany;
        {\tt\small mosbach@ais.uni-bonn.de}}%
}

\makeatletter
\let\@oldmaketitle\@maketitle
\renewcommand{\@maketitle}{\@oldmaketitle
      \begin{center}\vspace*{1ex}
      \includegraphics[width=\textwidth]{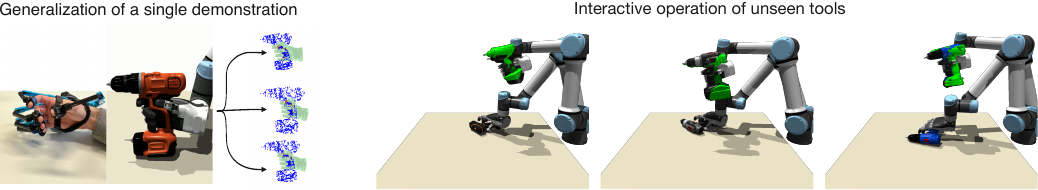}
      \label{overview}
      \vspace{-0.5cm}
      \captionof{figure}{A single grasping demonstration is transferred to other instances of a class, including instances not in the training set and only partially observed (left). 
      These generalized demonstrations guide the learning of an interactive policy able to operate a variety of tools (right).}
      \end{center}
      \vspace{-0.5cm}
    }
\makeatother

\begin{document}


\maketitle
\thispagestyle{empty}
\pagestyle{empty}

\renewcommand\thefigure{\arabic{figure}}    
\setcounter{figure}{1}

\begin{abstract}
Tool use, a hallmark feature of human intelligence, remains a challenging problem in robotics due the complex contacts and high-dimensional action space.
In this work, we present a novel method to enable reinforcement learning of tool use behaviors. 
Our approach provides a scalable way to learn the operation of tools in a new category using only a single demonstration. 
To this end, we propose a new method for generalizing grasping configurations of multi-fingered robotic hands to novel objects. 
This is used to guide the policy search via favorable initializations and a shaped reward signal.
The learned policies solve complex tool use tasks and generalize to unseen tools at test time.
Visualizations and videos of the trained policies are available at {\color{Blue} \url{https://maltemosbach.github.io/generalizable_tool_use}}.
\end{abstract}

\section{Introduction}
The use of tools to achieve desired changes to the environment is a hallmark feature of human intelligence~\cite{Jamone2022, Wenke2019}.
This includes various behaviors, from using a vessel to carry water, to driving a nail with a hammer, to operating a power drill. 
While humans routinely use specialized tools for construction and assembly tasks, this behavior has been challenging to automate because of the high-dimensional action space of humanoids robots and the intra-class variability of the tools made for human hands.

Despite these challenges, tool use remains a central task in robot learning, due to its overwhelming practical utility.
Classical approaches for tool use include affordance learning~\cite{Stoytchev2005, Montesano2007} and dynamic motion primitives~\cite{Kober2008, Muelling2010}.
These rely on predefined exploration primitives or trajectories and lack the interactive manipulation capabilities that humans so effortlessly exhibit.
Reinforcement learning (RL)~\cite{Rajeswaran2018,Wenke2019} has recently been used to generate interactive control policies, but suffers from the high-dimensional action space of human-like robotic hands. 
This leads to excessive sample complexity, if convergences can be achieved at all.
To enable RL to handle manipulation tasks in the intricacy of interactive tool use, auxiliary guidance such as demonstration datasets or precise reward engineering is needed~\cite{Rajeswaran2018}.

Using demonstrations to communicate the desired behavior to a robot is a intuitive approach since humans can provide competent demonstrations for anthropomorphic end-effectors. 
However, existing methods make only limited use of demonstrations.
Consider the task of grasping a hammer to drive a nail. 
To derive this general skill, regular imitation learning would necessitate a vast number of demonstrations spanning various tool instances.
Instead, we want our robot to use tools as flexibly as humans do, relating demonstrated behaviors to different instances without the need for repeated demonstrations.
While prior work considers intra-class variation of object instances~\cite{Rodriguez2020, Rodriguez2018}, the grasping of tools is framed as reaching a desired grasping position derived from an initial observation of the tool. 
This is in stark contrast to the way humans interact with tools and objects, where perception and action are continuously interleaved, making adaptive behaviors and operation in unstructured environments possible. 
Humans have the ability to effortlessly generalize prior knowledge and interactively adapt their behavior, enabling them to operate unfamiliar tools with ease.
While generalization to new tools and their interactive operation have been demonstrated individually, to the best of our knowledge, no prior method realizes both.

In this work, we present a system that learns a continuous control policy to operate a variety of tools under the guidance of only a single human demonstration. 
To this end, we introduce a procedure that utilizes non-rigid registration to generalize a canonical grasping demonstration to novel instances and use these demonstrations to guide policy search, without imposing rigid behaviors. 
This is achieved by initializing episodes in pre-grasp poses to enable efficient exploration and by inducing prior knowledge about how to grasp a tool through a shaped reward function.
Teh effectiveness of our proposed approach is experimentally evaluated on three simulated tool-use tasks.
Specifically, we make the following contributions:
\begin{itemize}
        \item We present a novel method that uses non-rigid registration to generalize grasp-configurations to unknown instances of a class.
        \item We examine how grasping demonstrations can be used to guide the learning of an RL policy.
        \item We demonstrate, in simulation, that interactive operation of different tools can be learned with model-free RL using only a single demonstration.
\end{itemize}

\section{Generalizing Demonstrated Grasps}

\begin{figure}[t]
        \centering
        \begin{subfigure}[b]{0.48\columnwidth}
                \centering
                \includegraphics[width=\textwidth]{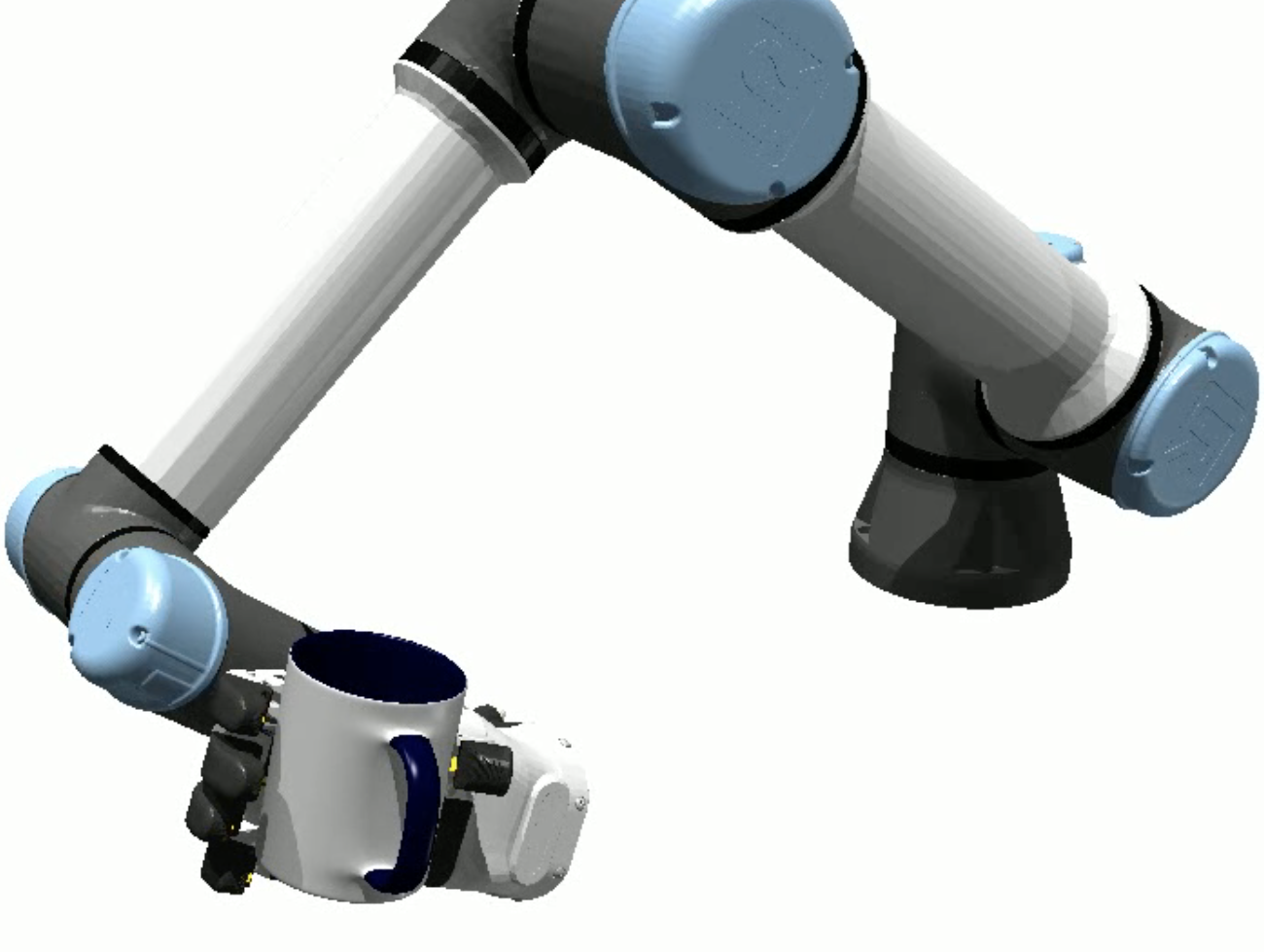}
        \end{subfigure}
        \hfill
        \begin{subfigure}[b]{0.48\columnwidth}
                \centering
                \includegraphics[width=\textwidth]{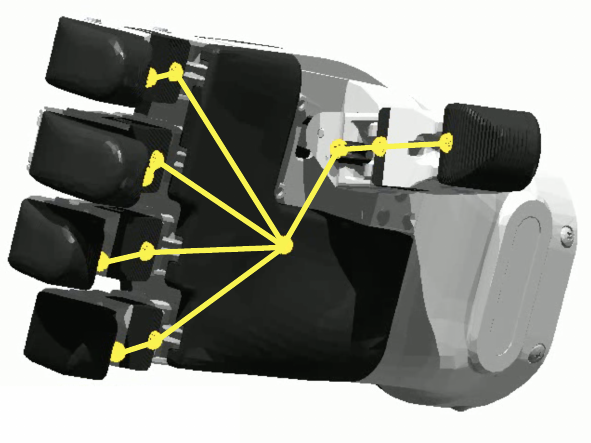}
        \end{subfigure}
        \caption{Task space of human-like grasping. Grasping and manipulating instances happens by making contact with the elements of the fingers and palm (left). 
        Hence, this naturally defines the task-space for multi-fingered robotic hands (right).}
        \label{fig:task_space_vectors}
\end{figure}

In grasping and tool use tasks, human-like robotic hands manipulate objects by inducing contact with the inside of the finger phalanges and the palm.
The corresponding keypoints, also referred to as task-space vectors~\cite{Qin2022a}, which are shown in Fig.~\ref{fig:task_space_vectors}, define the features of a grasp to be preserved during generalization to novel instances.
We found that such detailed multi-fingered grasping configurations can be accurately mapped between instances in a \textit{two-step approach}.
Specifically, we uniquely combine non-rigid registration and hand pose retargeting to construct a system for generalization of multi-fingered grasping configurations.
First, in Sec.~\ref{subsec:latent_deformation_field_manifold}, we leverage latent non-rigid registration to continuously deform the canonical object (and its demonstration keypoints) to match the observed object. 
This preserves characteristic category-level features of a grasp and works directly from partial point cloud observations. 
Second, in Sec.~\ref{subsec:optimization_in_task-space}, we optimize the end-effector pose and joint positions of the robot hand to find a kinematically feasible grasp that minimizes the distance in task space.

\subsection{Category-level Grasp Pose Transfer}
\label{subsec:latent_deformation_field_manifold}

\subsubsection{Coherent Point Drift}
To explain the first step in our grasp pose generalization method, we briefly review the coherent point drift (CPD) algorithm~\cite{Myronenko2010}.
Given a target point set $\bm{X} = (\bm{x}_1, \dots, \bm{x}_N)$ and a source point set $\bm{Y} = (\bm{y}_1, \dots, \bm{y}_M)$, our goal is to find a transformation that maps $\bm{Y}$ to $\bm{X}$.
CPD builds a Gaussian mixture model (GMM) from the moving point set, $\bm{Y}$, and treats the points in $\bm{X}$ as observations drawn from it.
An expectation-maximization (EM) algorithm  is used to optimize the GMM while obeying a smoothness constraint based on motion coherence theory~\cite{Yuille1988}.
The non-rigid transformation $\mathcal{T}$ mapping $\bm{Y}$ to $\bm{X}$ is given by: 
\begin{equation}
        \mathcal{T}(\bm{Y}, v) = \bm{Y} + v(\bm{Y}),
\end{equation}
where the displacement function $v$ is defined for any set of points $\bm{Z}$ as:
\begin{equation}
        v(\bm{Z}) = G(\bm{Y}, \bm{Z}) \bm{W}.
\end{equation}
$G(\cdot, \cdot)$ denotes a Gaussian kernel matrix. CPD estimates the matrix of kernel weights, $\bm{W}$, which can be understood as a set of deformation vectors associated with the points in $\bm{Y}$. 
Thus, the transformation from a canonical model $\bm{C}$ to a training instance $\bm{T}_i$ is defined as:
\begin{equation}
        \mathcal{T}_i(\bm{C}, \bm{W}_i) = \bm{C} + G(\bm{C}, \bm{C}) \bm{W}_i.
\end{equation}

\begin{figure}[b]
        \centering
        \rotatebox[origin=l]{90}{\makebox[0.0cm]{\hspace{-0.1cm} Observed \hspace{1.05cm} Canonical}}
        \begin{minipage}{.35\linewidth}
            \begin{subfigure}[t]{.9\linewidth}
                \includegraphics[width=2.25cm, height=2.25cm]{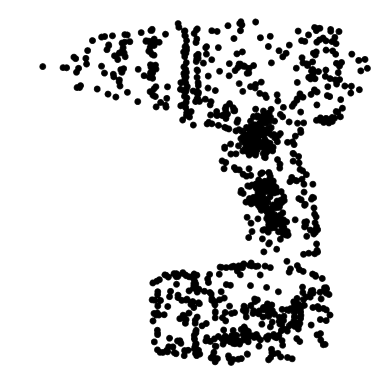}
            \end{subfigure} \\

            \begin{subfigure}[b]{.9\linewidth}
                \includegraphics[width=2.25cm, height=2.25cm]{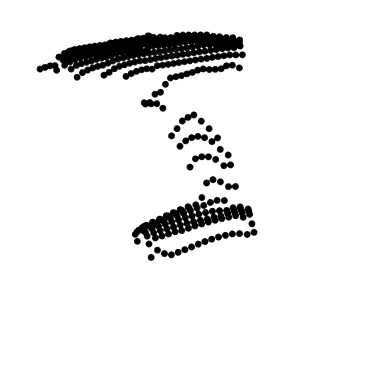}
            \end{subfigure} 
        \end{minipage}
        \hspace{0.01cm}
        \begin{minipage}{.57\linewidth}
            \begin{subfigure}[t]{1.01\linewidth}
                \includegraphics[width=\textwidth, height=\textwidth]{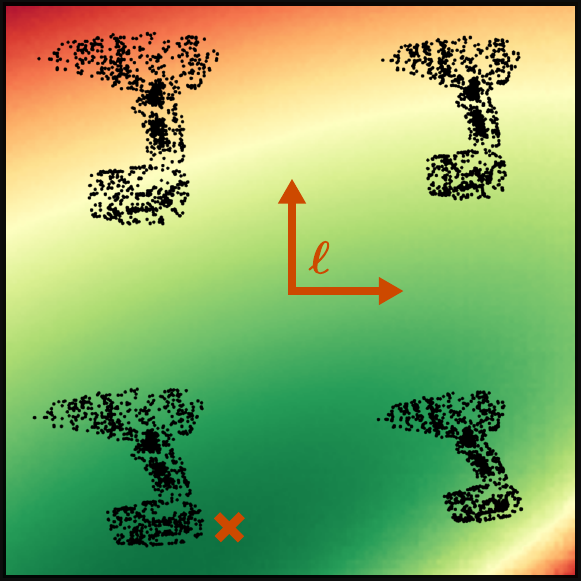}
            \end{subfigure}
        \end{minipage}
        \caption{Energy landscape induced by observed instance. 
        A partially observed instance induces an energy function over the latent shape parameters $\bm{\ell}$. 
        Optimizing this energy function yields latent space parameters that best map the canonical instance to the observation.}
        \label{fig:energy_landscape}
\end{figure}

\subsubsection{Latent Deformation Field Manifold}
Our goal is to extrapolate from a single demonstration to novel objects of the same category, utilizing understanding of common intra-class variability.
Therefore, we use CPD to find the deformation $\mathcal{T}_i(\bm{C}, \bm{W}_i)$ from the canonical instance, $\bm{C}$, to all other training instances $\bm{T}_i$.
The uniqueness of each deformation is captured entirely by $\bm{W}_i \in \mathbb{R}^{M \times 3}$. 
The corresponding row vectors $\bm{x}_i \in \mathbb{R}^{3M}$, which are the feature descriptors of the deformation fields, are assembled into a design matrix $\bm{X} \in \mathbb{R}^{n \times 3M}$, where $n$ is the number of training instances.
Finally, we perform principal component analysis (PCA) on $\bm{X}$, to find a lower-dimensional manifold of characteristic deformations $\bm{L} \in \mathbb{R}^{q \times 3M}$, where $q \ll n$ is the number of eigenvectors to keep, i.e., the dimensionality of our ensuing latent space.
Characteristic deformations of a category can now be modeled in the $q$-dimensional latent space.

\begin{figure}[t]
        \vspace{0.15cm}
        \centering
        \begin{subfigure}[b]{0.1875\columnwidth}
                \centering
                \includegraphics[width=\textwidth]{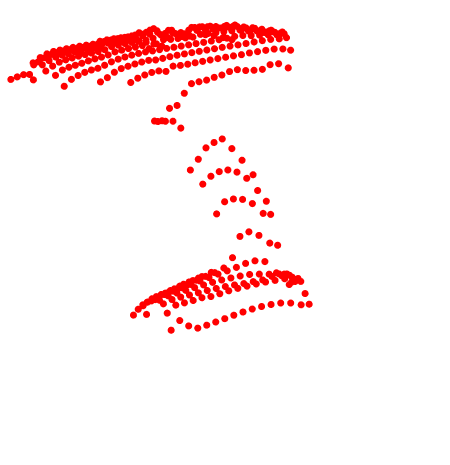}
        \end{subfigure}
        \hfill
        \begin{subfigure}[b]{0.1875\columnwidth}
                \centering
                \includegraphics[width=\textwidth]{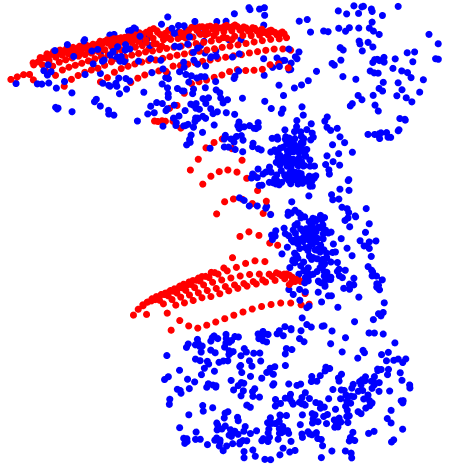}
        \end{subfigure}
        \hfill
        \begin{subfigure}[b]{0.1875\columnwidth}
                \centering
                \includegraphics[width=\textwidth]{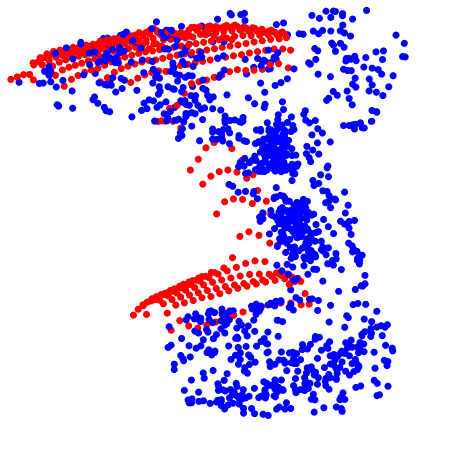}
        \end{subfigure}
        \hfill
        \begin{subfigure}[b]{0.1875\columnwidth}
                \centering
                \includegraphics[width=\textwidth]{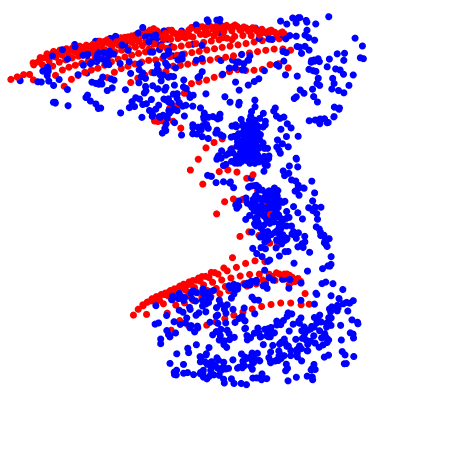}
        \end{subfigure}
        \hfill
        \begin{subfigure}[b]{0.1875\columnwidth}
                \centering
                \includegraphics[width=\textwidth]{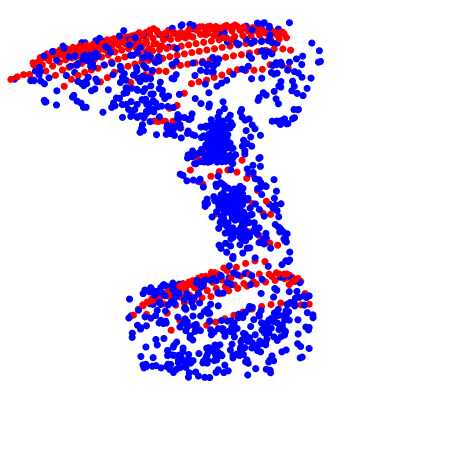}
        \end{subfigure}
        \caption{Registration of observed instance. The latent space parameters are optimized to fit the canonical instance (blue) to the observation (red).}
        \label{fig:register_observed}
\end{figure}

When encountering a new instance that has been partially observed through a segmented point cloud $\bm{O}$, we fit the latent parameter vector $\bm{\ell}$ to match its shape. Specifically, we aim to minimize the energy function:
\begin{equation}
        E(\bm{\ell}) = - \sum_{m=1}^{M} \log \sum_{n=1}^{N} e^{- \frac{1}{2\sigma^2} \lVert \bm{O}_n - \mathcal{T}(\bm{C}_m, W_m(\bm{\ell})) \rVert ^2},
				\label{eq:energy}
\end{equation}
illustrated in Fig.~\ref{fig:energy_landscape}, via gradient descent.  
Fig.~\ref{fig:register_observed} shows how this optimization deforms the canonical object to match the observed instance.
Since non-rigid registration in latent space permits only deformations actually observed in a class, we can register even partially observed objects without invoking undesired deformations of the canonical model.
Once a minimum is found, we can use the resulting deformation field to transform the keypoints of the canonical demonstration into generalized keypoint poses.

\subsection{Pose Regression in Task Space}
\label{subsec:optimization_in_task-space}

\begin{figure}[t]
        \centering
        \includegraphics[width=\columnwidth]{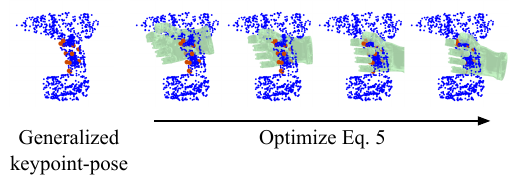}
        \caption{Pose regression in task-space. Given a generalized keypoint-pose (left), an attainable grasp pose is found by minimizing the keypoint distance (right).}
        \label{fig:2_step_generalization}
\end{figure}

So far, we have only shown how the deformation field between canonical and observed instances can be used to transform feature points of a grasp.
While previous work~\cite{Stuckler2014, Rodriguez2018} uses this property to generalize trajectory control poses, 
this forgoes the inherent advantage of multi-fingered manipulators to grasp in diverse finger configurations. 
Our aim is to maintain the properties of a grasp when it is transferred to new objects, which entails determining a relationship between the deformation of an object and the joint angles of the resultant grasp.
Thus, while the transferred keypoints represent the desired hand pose, they might not define a reachable position under the kinematic constraints of the used end-effector.  
To address this issue, we introduce a second optimization step, inspired by motion retargeting approaches~\cite{Qin2022,Qin2022a}, to find an attainable grasp configuration.
We optimize the wrist pose $\bm{p}$ and joint positions $\bm{q}$ of the robot hand to minimize the distance to the transferred keypoints $\bm{k}_i$:
\begin{equation}
        \min_{\bm{p}, \bm{q}} \sum_{i=0}^{N} \lVert \bm{k}_i - f_i(\bm{p}, \bm{q}) \rVert ^2,
        \label{eq:task_space_objective}
\end{equation}
where $f_i$ represents the forward kinematics of the $i$th keypoint. 
Solving this optimization problem yields the hand pose and joint angles that best preserve task-space characteristics of the demonstrated grasp.
Fig.~\ref{fig:2_step_generalization} shows the process of optimizing for minimal task-space distance (Eq.~\ref{eq:task_space_objective}).
The robot hand converges to a feasible grasping position that maintains characteristic features of the original demonstration.

\section{Interactive Tool Use}
Thus far, we have introduced our method for intra-class generalization of functional grasps. 
However, we only concerned ourselves with static grasp poses.
In the following, we describe how the obtained demonstrations can be used to guide the learning process of interactive RL policies.
Sec.~\ref{subsec:pre-grasp_guided_exploration} presents pre-grasp poses as efficient exploration primitives.
In Sec.~\ref{subsec:rl_with_privileged_information}, we propose a shaped reward function to direct the policy based on the generalized grasp poses.
Lastly, Sec.~\ref{subsec:transfer_to_unseen_tools} discusses how grasp poses and tool-use policies can generalize to instances beyond the training set.

\subsection{Pre-grasp Poses for Efficient Exploration}
\label{subsec:pre-grasp_guided_exploration}
The process of grasping tools can be described by an initial \textit{reaching} and a subsequent \textit{dexterous manipulation} phase~\cite{Dasari2023}.
The first phase, where the robot reaches for the tool but does not yet make contact, can be solved very efficiently by conventional feedback control methods.
Only the subsequent high-contact manipulation requires an interactive RL policy.
In this context, our generalized grasp poses can be utilized to favorably position the robotic hand at the beginning of each episode by moving to a pose offset from our final target.
Specifically, we have the robot approach a pose that is removed from the target grasp in the direction normal to the palm, and interpolate the joint angles between the open and target configurations.
On the right of Fig.~1, we overlay both the pre-grasp configuration, which represents the beginning of the episode, and a later configuration just before the task is completed. 
Pre-grasp poses serve as a critical precursor to efficient exploration and successful learning of the task at hand.

\begin{table*}[t]
        \caption{Grasp-pose guided reward. The reward function combines {\color{dark_green}\bf task-specific rewards} encouraging goal-directed behavior and {\color{dark_orange}\bf tool grasping rewards} that encourage the agent to reach the demonstrated grasping pose. }
        \begin{minipage}[b]{1 \linewidth}
          \centering 
          {
          \small
          \begin{tabular}{llr}
          \toprule
          Term & Equation & Weight \\
          \midrule
          \tabnode{\phantom{0}}\hspace*{-0.6ex}$\circ$\,\textit{Place mug} & & \vspace*{-1ex} \tabnode{\phantom{$0.0$}}  \\
          \phantom{0}~~~$r_\textrm{pose}$: target pose matching & $e^{-\alpha \lVert x_t^{(p)} - \overline{x}_t^{(p)}\rVert - \beta \angle (x_t^{(o)}, \overline{x}_t^{(o)})}$ & $25.0$ \\
          \phantom{0}~~~$r_\textrm{success}$: target pose reached & $\mathds{1}(\textrm{pose\_reached})$ & \vspace*{0.5ex} $100.0$ \\
          \phantom{0}$\circ$\,\textit{Position drill} & &  \vspace*{-1ex} \\
          \phantom{0}~~~$r_\textrm{pose}$: target pose matching & $e^{-\alpha \lVert x_t^{(p)} - \overline{x}_t^{(p)}\rVert - \beta \angle (x_t^{(o)}, \overline{x}_t^{(o)})}$ & $25.0$ \\
          \phantom{0}~~~$r_\textrm{success}$: target pose reached & $\mathds{1}(\textrm{pose\_reached})$ & \vspace*{0.5ex} $100.0$ \\
          \phantom{0}$\circ$\,\textit{Drive nail} & &  \vspace*{-0.5ex} \\
          \phantom{0}~~~$r_\textrm{dist}$: move hammer to nail & ${(\epsilon + \Delta x)}^{-1}$ & $0.25$ \\
          \tabnode{\phantom{0}}~~$r_\textrm{depth}$: nail depth & $\Delta d_\textrm{nail}$ & \vspace*{0.5ex} \tabnode{$100.0$ } \\
          
          \tabnode{\phantom{0}}~~$r_\textrm{kp}$: keypoint matching & ${(\epsilon + \Delta k)}^{-1}$ & \tabnode{$0.001$ } \\
          \tabnode{\phantom{0}}~~$r_\textrm{lift}$: tool lifting& ${(\epsilon + \Delta h)}^{-1}$ & \tabnode{$0.05$ } \\
           \bottomrule
          \end{tabular}
          }
          
          \label{tbl:shaped_reward}
          \end{minipage}
          
          \begin{tikzpicture}[overlay]
          \draw [dark_green] (1.north west) -- (2.north east) --
          (4.south east) -- (3.south west) -- cycle;
          \draw [dark_orange] (5.north west) -- (6.north east) --
          (8.south east) -- (7.south west) -- cycle;
          
          \node [right=2cm,below=0.7cm,minimum width=0pt,anchor=west] at (2) (A) {{\color{dark_green}{\small \bf Task-specific}}};
          \draw [<-,out=0,in=180] (2) to (A);
          \node [right=2cm,above=0.6cm,minimum width=0pt,anchor=west] at (6) (B) {{\color{dark_orange}{\small \bf Tool grasping}}};
          \draw [<-,out=0,in=180] (6) to (B);
          \node [right=0.8cm,above=0.1cm,minimum width=0pt,anchor=west] at (8) (C) {{\color{black}{\footnotesize We use $\alpha=10$, $\beta=1$,}}};
          \node [right=0.8cm,below=0.2cm,minimum width=0pt,anchor=west] at (8) (C) {{\color{black}{\footnotesize  and $\epsilon=0.025$.}}};
          \end{tikzpicture}
      \vspace{-0.3cm}
			
\end{table*}

\subsection{Grasp-pose Guided RL}
\label{subsec:rl_with_privileged_information}
The operation of various tools can be mastered efficiently once robust grasps have been learned.
Hence, the generalized grasp poses are used to inject prior knowledge on how a tool ought to be grasped by parametrizing a shaped reward function.
The reward terms encouraging the agent to reach the demonstrated grasp-pose are detailed at the bottom of Tab.~\ref{tbl:shaped_reward}.
Specifically, an incentive is provided to minimize the distance to the demonstrated keypoints. 
A second reward term trains the agent to pick up the tool from the table, which requires learning a stable grasp and simple maneuvering of the tool.
We have found that defining the object height in terms of the object's root coordinate system can lead to undesirable local optima. 
For example, the coordinate root of the drill is at the tip of the tool, which causes the agent to learn unhelpful solutions, such as tilting the drill up without ever actually lifting it.
To avoid these problems, we decided to sample a synthetic point cloud on the tool's mesh. 
The height of the tool is then defined by the height of its lowest surface point, which is an approximation of the actual clearance between the tool and the table.
We use proximal policy optimization~\cite{Schulman2017} to train our policies to maximize this reward.

\subsection{Transfer to Unseen Tools}
\label{subsec:transfer_to_unseen_tools}
Ultimately, we want the learned policies to be able to operate unseen tools.
Thus, our goal is to generalize the grasp pose to a novel instance without access to its object mesh.
We therefore add a depth camera sensor to the environment, as shown in Fig.~\ref{fig:visual_environment_setup}. 
Unlike the synthetic point clouds, the sensor data suffers from occlusions.
Simply applying CPD would now deform the canonical instance in unhelpful ways. 
However, the learned category-level shape space can circumvent this problem. 
Since the low-dimensional deformation field manifold only allows for deformations that are characteristic of the variance in a class, the canonical model can even be fitted to a partially observed instance.

\begin{figure}[b]
        \centering
        \begin{subfigure}[b]{0.32\columnwidth}
                \centering
                \includegraphics[width=\textwidth]{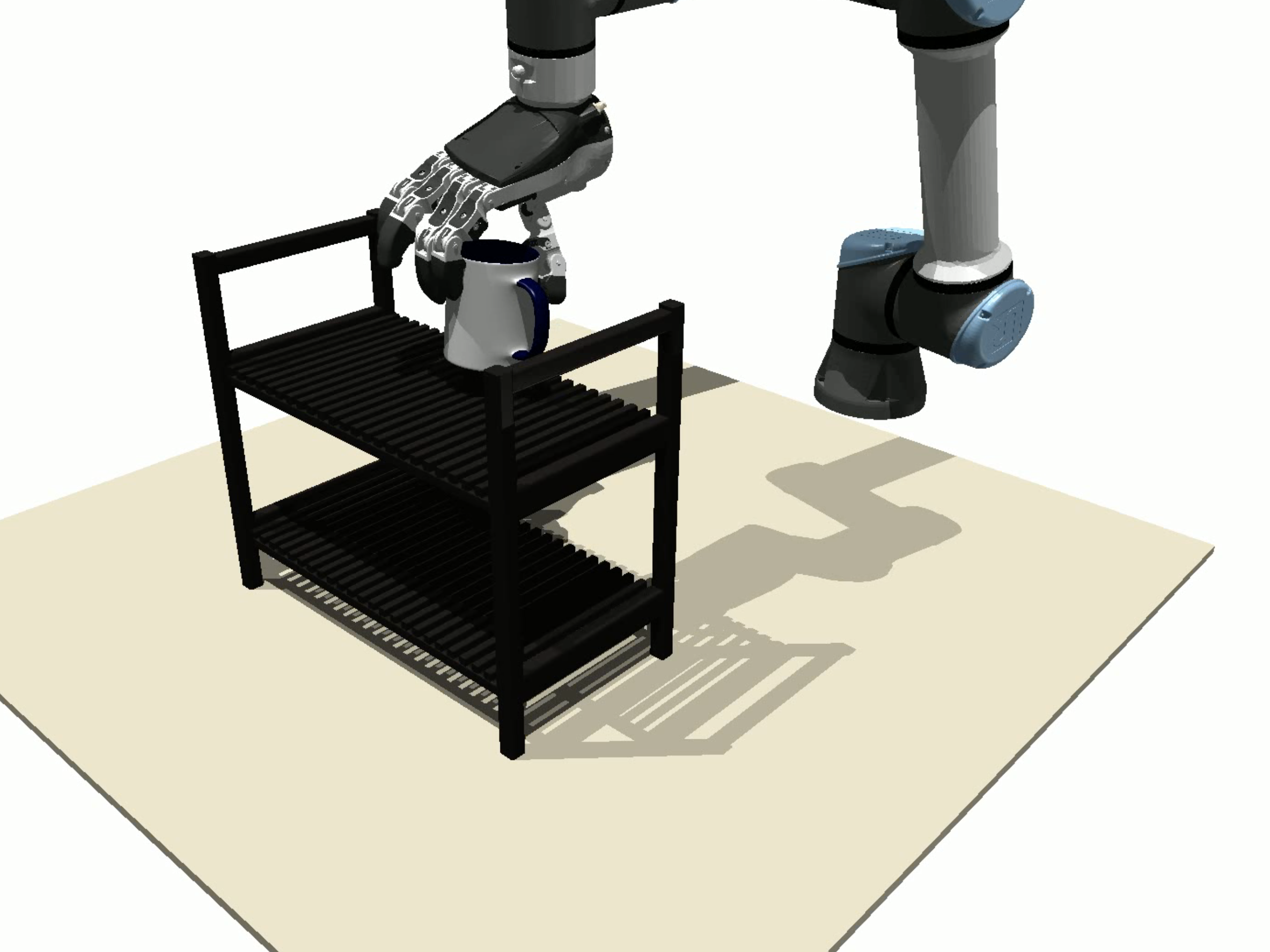}
                \subcaption*{Place mug}
        \end{subfigure}
        \hfill
        \begin{subfigure}[b]{0.32\columnwidth}
                \centering
                \includegraphics[width=\textwidth]{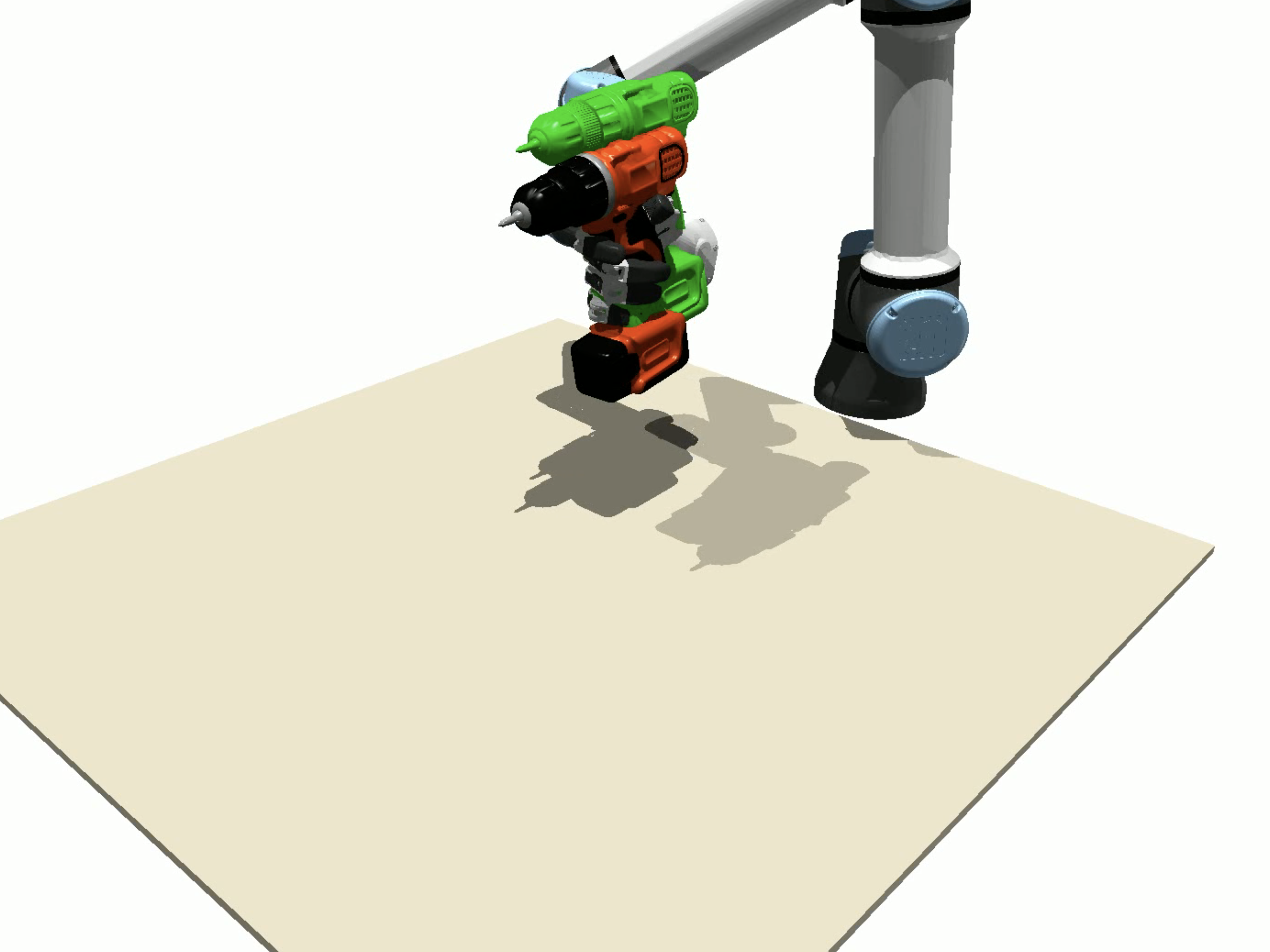}
                \subcaption*{Position drill}
        \end{subfigure}
        \hfill
        \begin{subfigure}[b]{0.32\columnwidth}
                \centering
                \includegraphics[width=\textwidth]{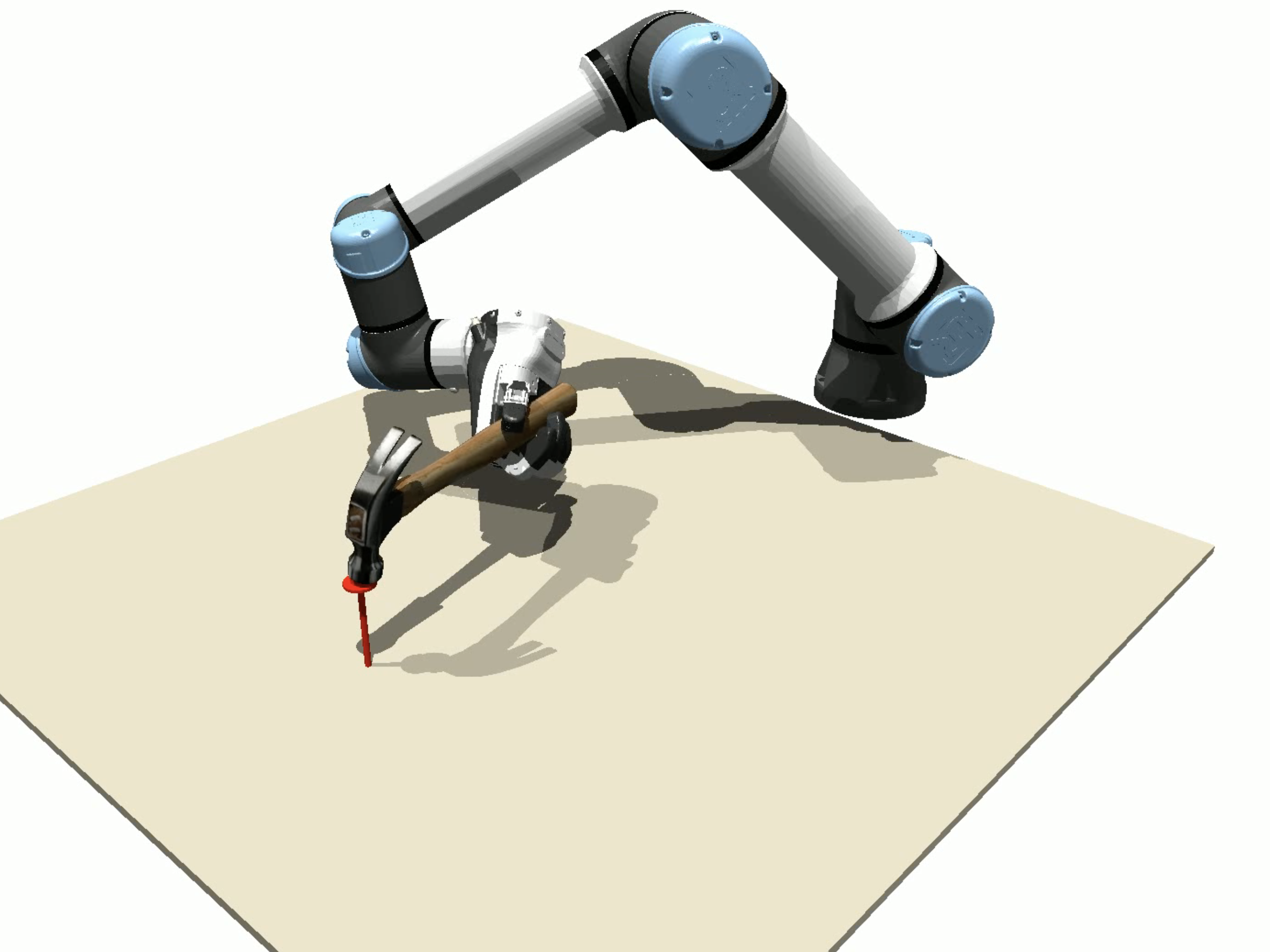}
                \subcaption*{Drive nail}
        \end{subfigure}
        \caption{Tool use tasks. The environments represent familiar tool use tasks a robot might be asked to solve.}
        \label{fig:task_overview}
\end{figure}

\begin{figure}[b]
        \includegraphics[width=\columnwidth]{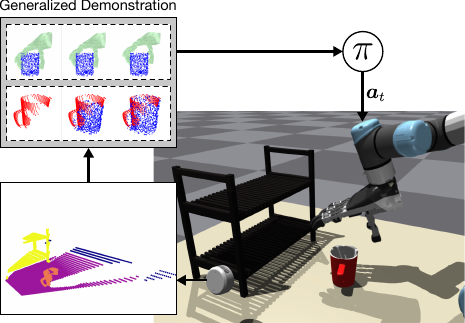}
        \caption{Grasp generalization from vision. We add a sensor to the simulation that outputs segmented point clouds. We fit the canonical model of the respective object category to the measurements belonging to the tool (Eq.~\ref{eq:energy}). We then find the joint configuration that minimizes the task-space distance (Eq.~\ref{eq:task_space_objective}) and pass this generalized demonstration to our policy.}
        \label{fig:visual_environment_setup}
\end{figure}

\section{Experimental Setup}
Our experiments aim to answer how effectively the proposed method can solve the challenging task of robotic tool use based on a single demonstration.
Specifically, we evaluate
(1) Whether a canonical demonstration can be generalized to new instances; 
(2) how effectively model-free RL can solve the posed tasks based on the generalized demonstrations; 
(3) whether our policies can generalize to novel, partially observed tools in a zero-shot manner.

\subsection{Problem Statement}
Our goal is to learn a policy $\pi$ to utilize a tool (${T_i | i=1, \dots, N}$) in order to achieve some goal-directed behavior, e.g. using a hammer to drive a nail. 
Moreover, the policy should be able to operate a variety of tool instances and generalize to unseen tools at test time. 
In a repeated interaction, the policy observes the current state of the environment $\bm{s}_t \in \mathcal{S}$, performs an action $\bm{a}_t \in \mathcal{A}$, and receives a reward signal $r_t$.
We define the observations of the policy to include proprioceptive observations of the robot state (wrist pose and keypoints of the hand), as well as a low-dimensional observation of the tool represented by its generalized demonstration and latent shape parameters. 
Additionally, the policy receives information about task-specific objectives, such as the desired pose of the drill.
The action space $\mathcal{A}$ comprises the desired change to the end-effector pose and joint positions of the robot hand. The agent chooses actions at a frequency of 30\,Hz.
The reward function is the sum of the terms detailed in Tab.~\ref{tbl:shaped_reward}.

\subsection{Tool Use Tasks}
We evaluated our method on three tool categories: \textit{Drills}, \textit{Hammers}, and \textit{Mugs}, each with one canonical, 10 training, and 3 test instances. 
The models were obtained from the online databases GrabCAD\footnote{\url{https://grabcad.com/library}}, 3DWarehouse\footnote{\url{https://3dwarehouse.sketchup.com/}}, and Sketchfab\footnote{\url{https://sketchfab.com}}.
The simulated robot combines a UR5e arm controlled by its end effector pose with a Schunk SIH hand that has 11 degrees of freedom (DoFs), 5 of which are fully actuated.
We use NVIDIA Isaac Gym~\cite{Makoviychuk2021} to simulate the tool use tasks shown in Fig.~\ref{fig:task_overview}. 
In each run, 16,384 parallel agents are trained for a total of 134 million simulated steps, which corresponds to approximately 52 real-time days. 
This requires just under 3 hours of wall-clock time on a single NVIDIA A6000 GPU. At test time, we required an average of 3 seconds to match the canonical model to an observed instance.

\subsection{Demonstrations}
Our method draws on human grasping knowledge to accelerate the learning process. 
To demonstrate grasping postures in an intuitive way, we introduce a virtual reality (VR) interface to Isaac Gym. 
The operator's movements are tracked by a SenseGlove DK1, which captures finger angles, and an HTC Vive tracker, which records the hand pose.
This device, worn by the operator, can be seen on the left in Figure~1.
An HTC Vive headset is integrated with Isaac Gym's camera sensors to provide a stereoscopic visualization of the scene.
The operator interacts with the tasks in a natural way, indicating at the push of a button that the current pose should serve as the canonical demonstration.

\subsection{Evaluation Procedure}
For the \textit{Place mug} and \textit{Position drill} tasks, the success criterion is based on the distance of the tool pose and target pose. 
We consider an episode as completed successfully if $d < \bar{d}$ and $\theta < \bar{\theta}$, where $d$ and $\theta$ are the positional and angular distance to the target pose.
For both environments we use $\bar{d} = 0.03\mathrm{m}$ and $\bar{\theta} = 0.2\mathrm{rad}$.
The \textit{Drive nail} environment considers runs successful, where the nail has been driven by a depth of greater than $0.075\mathrm{m}$.

\begin{table}
        \caption{Task-space distance.}
        \centering \normalsize
        \begin{tabular}{rrrr} \toprule
                & Mugs & Drills & Hammers  \\ \midrule
                Ours  & $\mathbf{0.68 \pm 0.26}$ & $\mathbf{0.72 \pm 0.30}$ & $\mathbf{0.78 \pm 0.34}$ \\
                WP  & $2.27 \pm 1.12$ & $2.76 \pm 1.74$ & $2.64 \pm 1.12$ \\
                CG  & $2.00 \pm 0.97$ & $2.89 \pm 1.45$ & $3.88 \pm 2.22$ \\ \bottomrule \vspace*{-3mm}
        \end{tabular}

				\footnotesize Mean distance in cm of the grasps proposed by our method and\\ ablations to the keypoints of the generalized demonstration.
        
        \label{tbl:task_space_distance}
\end{table}

\section{Results}

\subsection{Analysis of Generated Grasps}
\begin{figure}[t]
        \centering
        \rotatebox[origin=l]{0}{\makebox[0.0cm]{\hspace{-0.8cm} Demonstration \hspace{0.75cm} Ours \hspace{1.3cm} WP \hspace{1.3cm} CG}}

        \vspace{0.1cm}
        \begin{minipage}{.25\linewidth}
                \begin{subfigure}[t]{\linewidth}
                    \includegraphics[width=1.0\textwidth, height=0.75\textwidth]{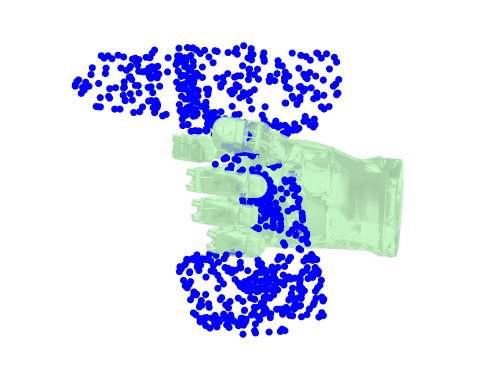}
                \end{subfigure}
        \end{minipage}
        \hspace{0.2cm}
        \begin{minipage}{.7\linewidth}
            \begin{subfigure}{.31\columnwidth}
                \includegraphics[width=\textwidth, height=0.75\textwidth]{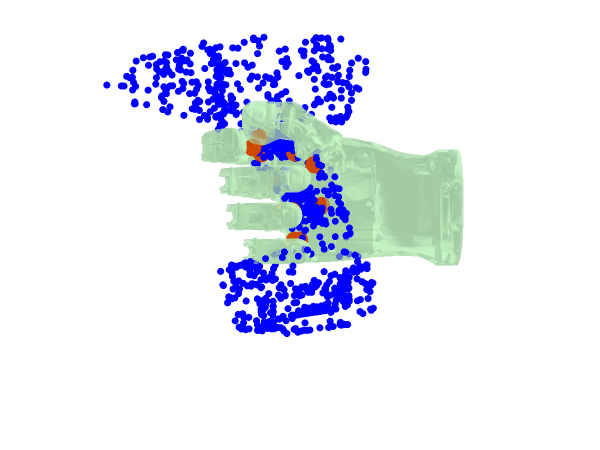}
            \end{subfigure}
            \hfill
            \begin{subfigure}{.31\columnwidth}
                \includegraphics[width=\textwidth, height=0.75\textwidth]{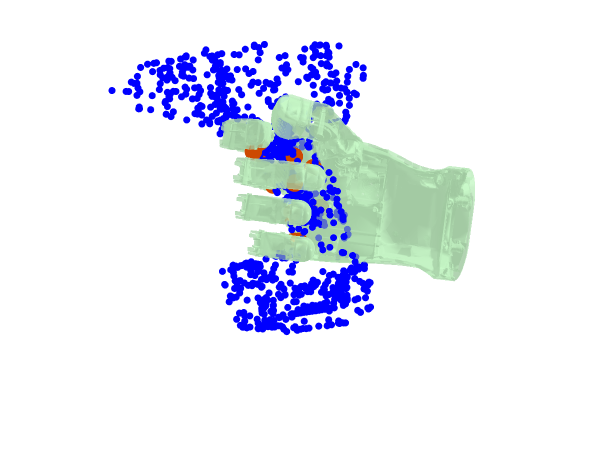}
            \end{subfigure}
            \hfill
            \begin{subfigure}{.31\columnwidth}
                \includegraphics[width=\textwidth, height=0.75\textwidth]{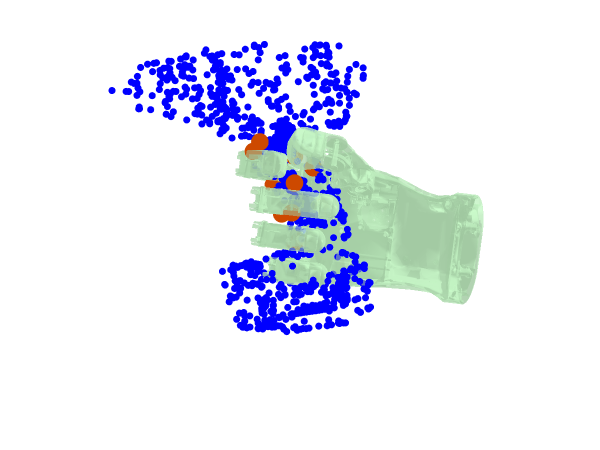}
            \end{subfigure}

            \vspace{0.1cm}
            \begin{subfigure}{.31\columnwidth}
                \includegraphics[width=\textwidth, height=0.75\textwidth]{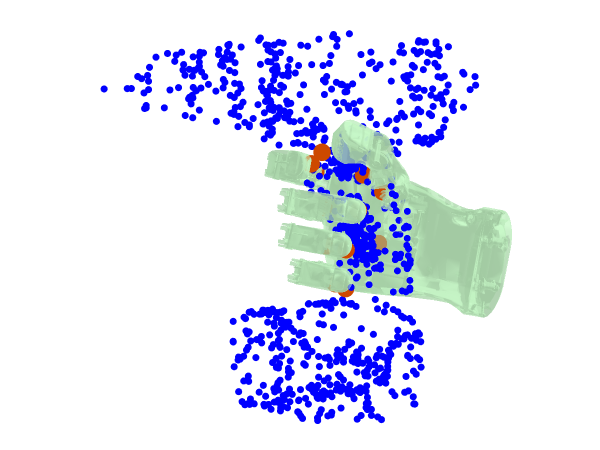}
            \end{subfigure}
            \hfill
            \begin{subfigure}{.31\columnwidth}
                \includegraphics[width=\textwidth, height=0.75\textwidth]{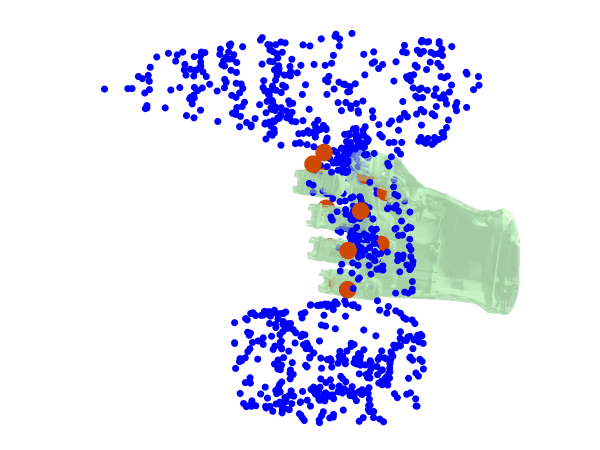}
            \end{subfigure}
            \hfill
            \begin{subfigure}{.31\columnwidth}
                \includegraphics[width=\textwidth, height=0.75\textwidth]{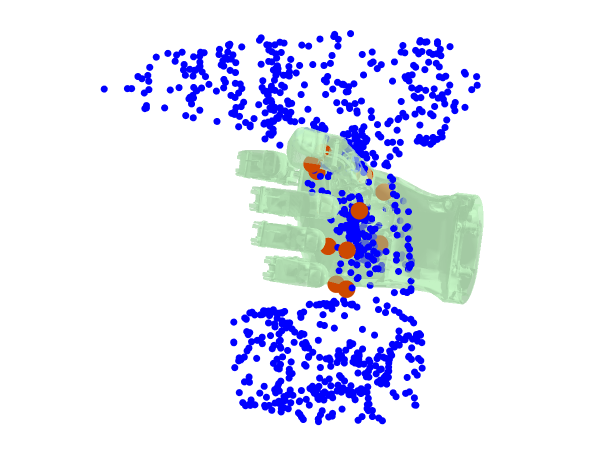}
            \end{subfigure}
        \end{minipage}
        
        \caption{Generated grasp-poses. The proposed approach generates grasp poses that aim to be equivalent in task-space. 
        Generalizing trajectory control poses, such as the wrist pose to be reached before closing the fingers (WP) does not have this desired property.}
        \label{fig:generated_grasp_poses}
\end{figure}

First, we investigate the kind of grasp poses that the proposed framework generates. 
Here we compare with two ablations: Retention of the canonical grasp (CG) and transformation of the wrist pose while keeping grasping behavior constant (WP).
To assess the quality of a grasp, we measure the distance to the transformed keypoints over all training instances in a class. 
Quantitative results are shown in Tab.~\ref{tbl:task_space_distance}.
The proposed method outperforms both baselines by a large margin, and the results are consistent across tool categories.
Examples of the generalized grasp-poses shown in Fig.~\ref{fig:generated_grasp_poses} confirm that our approach finds feasible grasps for varying object shapes.

\subsection{Grasp-pose Guided RL}
Next, we evaluate the ability of generalized grasp pose demonstrations to guide policy search on challenging tool use tasks. 
The results in Tab.~\ref{tbl:training_performance} show that the proposed method consistently finds the intended grasps across the tasks and tools studied. 
Furthermore, the Place Mug and Position Drill tasks are solved with high reliability. 
Driving a nail proved to be the most challenging task to complete, as the agent must maintain its grasp on the hammer while making forceful contact with the environment.
Now, we compare the performance of our proposed method to multiple ablations. 
First, we examine how performance changes when we disable our grasp generalization (w/o GG) and instead apply canonical demonstration to all objects.
As can be seen, this still leads to viable training performance for objects with lower variance, such as cups, while performance deteriorates more severely for objects that vary greatly in their extent and grasping position, such as drills and hammers.
Not navigating to a pre-grasp pose at the beginning of the episode (w/o PG) causes the training to fail.
The agent is not able to find the correct grasp posture, but frequently gets stuck in local optima.
Lastly, we compare to a baseline where the task is approached without demonstration guidance (w/o demo). 
Here, the agent receives only task-specific rewards and is initialized in a default neutral position above the table.
Again, learning the full manipulation tasks is unsuccessful, as discovering useful behaviors that make progress on the proposed task is extremely difficult in this situation.

The results show that knowledge about how to grasp an object, which can be incorporated via shaped rewards or pre-grasp poses, is a valuable addition to RL training.
Moreover, having a method that generalizes such demonstrations to new objects in a class removes the high overhead of collecting a large number of demonstrations and allows training to scale more easily.

\begin{table}
    \caption{Training performance. Success rates of the proposed method and studied ablations to grasp the tool and solve the full task.}
    \centering
    \begin{tabular}{rrrcrrcrr} \toprule
           & \multicolumn{2}{c}{Place mug} & \phantom{} & \multicolumn{2}{c}{Position drill} & \phantom{} & \multicolumn{2}{c}{Drive nail}  \\
           \cmidrule{2-3} \cmidrule{5-6} \cmidrule{8-9} 
         & grasp & full & \phantom{} & grasp & full & \phantom{} & grasp & full \\ \midrule

        Ours  & $\mathbf{0.97}$ & $\mathbf{0.96}$ & \phantom{} & $\mathbf{0.94}$ & $\mathbf{0.76}$ & \phantom{} & $\mathbf{0.8}$ & $\mathbf{0.65}$ \\
        w/o GG  & $\mathbf{0.97}$ & 0.95 & \phantom{} & 0.81 & 0.66 & \phantom{} & 0.74 & 0.61 \\
        w/o PG  & 0.01 & 0.0 & \phantom{} & 0.41 & 0.0 & \phantom{} & 0.4 & 0.0 \\
        w/o demo  & 0.0 & 0.0 & \phantom{} & 0.0 & 0.0 & \phantom{} & 0.0 & 0.0 \\ \bottomrule
    \end{tabular}
    \label{tbl:training_performance}
\end{table}

\subsection{Zero-shot Transfer to Unseen Tools}
Finally, we investigate whether the policy is able to transfer to unseen tools in a zero-shot manner.
Here, we do not assume access to the object mesh, instead perceiving the scene via a segmented point-cloud, as shown in Fig.~\ref{fig:visual_environment_setup}.
The canonical demonstration is then adjusted to fit the observed instance and given to the policy. 
It can be seen in Tab.~\ref{tbl:test_performance}, that the policies can grasp and operate even some of the unseen tools without finetuning. 
Extending the training set may help to close the performance gap between the training and test instances in the future.

\section{Related Work}
\subsubsection{Robotic grasping}
Despite decades of active research efforts, robotic grasping remains an unsolved problem~\cite{Zeng2022}.
Grasping has traditionally been framed as the open-loop procedure of grasp-pose prediction (grasp synthesis). 
Several prior works estimate grasp-poses through analytical~\cite{Sahbani2012,Ponce1993,Ding2000} or learned~\cite{Bohg2013,Kleeberger2020} methods.
In recent years, RL has become popular for robotic grasping and manipulation due to its ability to generate interactive policies in a model-free manner. 
Kalashnikov et al.~\cite{Kalashnikov2018} train a vision-based grasping policy to control a parallel gripper. 
Shahid et al.~\cite{Shahid2020} demonstrate that RL can be used to continuously control a Franka Emika Panda manipulator to lift objects off a table.
However, RL has struggled with the high-dimensional action space of anthropomorphic end-effectors. 
One group of work has aimed to scale up experience collection via parallelized GPU-accelerated physics simulation~\cite{Makoviychuk2021,Chen2022,Mosbach2022a}.
Alternatively, human demonstrations have been used by themselves~\cite{Qin2022,Qin2022a} or in combination with RL~\cite{Mosbach2022,Rajeswaran2018} to solve grasping and manipulation tasks.

\subsubsection{Robotic tool use}
Prior works studying robotic tool use span classical~\cite{Brown2013,Toussaint2018} and learning-based~\cite{Fang2020,Wenke2019, Xie2019} approaches. 
Xie et al.~\cite{Xie2019} learn to predict the visual outcome of actions based on human demonstration and autonomous interaction data.
Planning with the learned model can solve improvised tool use tasks with a parallel gripper.
Wenke et al.~\cite{Wenke2019} study reasoning and generalization in RL through the lens of tool use. 
They train RL agents to solve grid-world versions of the classical trap-tube experiment.
Notably, Dasari et al.~\cite{Dasari2023} demonstrate that pre-grasp poses can be used to improve dexterous manipulation learning. 
While their objective of using grasp-poses to accelerate RL is aligned with the goal of our work, they do not consider generalization of grasp-poses or policies between different tools.
To the best of our knowledge, the amalgamation of transferring demonstrations between instances and interactive RL training is novel to our work.

\begin{table}
        \caption{Test performance. Success rate of the proposed method when operating unseen tools. The generalized demonstration is estimated from a partial point-cloud of the tool.}
        \centering
        \begin{tabular}{rrcrcr} \toprule
               & \multicolumn{1}{c}{Place mug} & \phantom{} & \multicolumn{1}{c}{Position drill} & \phantom{} & \multicolumn{1}{c}{Drive nail}  \\ \midrule
              & $0.67 \pm 0.11$ & \phantom{} & $0.62 \pm 0.15$ & \phantom{} & $0.55 \pm 0.1$ \\ \bottomrule
        \end{tabular}
        \label{tbl:test_performance}
\end{table}

\subsubsection{Grasp-pose transfer.}
Multiple lines of work aim to generalize demonstrated behaviors to novel instances in a class. 
Object-meshes segmented via shape and volumetric information are used by Vahrenkamp et al.~\cite{Vahrenkamp2016} to transfer grasps from a template set to familiar objects.
Stückler et al.~\cite{Stuckler2014} transfer poses and trajectories defining grasping motion via the dense deformation field from the known object model to an observed instance.
Rodriguez et al.~\cite{Rodriguez2018} extend this work by modelling deformations not only between a known and observed instance, but within a category. 
This makes it possible to register partially observed objects.
In~\cite{Stouraitis2015} and~\cite{Amor2012}, contact points are warped from a known object to an observed object. 
However, both assume that the objects are fully observed.
Simeonovdu et al.~\cite{Simeonovdu2021} present neural descriptor fields which represent an object by a mapping from each 3D point $\bm{x}$ to a latent descriptor $z$ encoding relations to salient object features.
This description is used to establish correspondences of semantically meaningful object features, and thereby generalize demonstrations to new instances.
Our work builds on~\cite{Rodriguez2018}, but generalizes characteristic features of a multi-fingered grasp through optimization in task space. 
Further, we demonstrate how the generalized demonstrations can be used as a basis for learning interactive tool use policies, rather than as parameters for open loop grasping behavior.

\section{Discussion and Conclusion}
We have shown that the challenging domain of robotic tool use becomes approachable for model-free RL with the use of only a single human demonstration.
The proposed generalization scheme can transfer grasp poses even to partially observed instances while retaining characteristic features of the demonstrated functional grasp.
The RL experiments underscore the benefits of extending grasp pose generalization to the domain of interactive control, as the policies are for example able to continuously manipulate drills lying on the table until a desired grasp is achieved.
Although we only present results in simulation, we have shown how the latent shape parameters and grasping configuration of a novel object can be estimated from its partial point-cloud observation.
Still, there are several limitations and opportunities for future work. 
Transferring the obtained results to the real robot system is the most evident task. 
Developing a way to track tools during the grasping process and obtain well separated point-clouds of a scene are key challenges to be overcome.
In addition, developing an approach that can generate class-independent grasping or pre-grasp poses would be valuable.

\section*{Acknowledgement}
\small{This work has been funded by the German Ministry of Education and Research (BMBF), grant no. 01IS21080, project “Learn2Grasp: Learning Human-like Interactive Grasping based on Visual and Haptic Feedback”.}

\bibliographystyle{IEEEtran}
\bibliography{references.bib}

\end{document}